\documentclass{article}
\pdfoutput=1
\usepackage[preprint,nonatbib]{neurips_2020}
\usepackage{amssymb}
\usepackage{amsmath}
\usepackage{amsthm}
\theoremstyle{plain}
\usepackage{bbm}
\usepackage{mathtools}
\usepackage{bm}
\usepackage{csvsimple}
\usepackage{dirtytalk}
\usepackage{hyperref}
\usepackage{mdframed}
\author{
  Karim~Lounici \\
  C.M.A.P.\\
  Ecole Polytechnique\\
  91120 Palaiseau\\
   \And
    Katia~Meziani \\
  CEREMADE\\
  Paris Dauphine-PSL\\
  75016 Paris \\
 \And
   Benjamin~Riu\thanks{CIFRE convention with Uptilab.} \\
  C.M.A.P.\\
  Ecole Polytechnique\\
  91120 Palaiseau\\
}
\newcommand{\ER}{\textit{\texttt{ER}}}
\newcommand{\GER}{\textit{\texttt{ER-G}}}
\newcommand{\BKK}{\textit{\texttt{BKK}}}
\newcommand{\BKKs}{\textit{\texttt{BKKs}}}
\newcommand{\R}{\ensuremath{\mathbb{R}}}

\newcommand{\e}{\xi}
\newcommand{\be}{\beta}
\newcommand{\x}{\mathbf{x}}
\newcommand{\spa}{\gamma}
\newcommand{\LAS}{Lasso }

\newcommand{\Emu}{\mu}
\newcommand{\Sig}{\textbf{S}}
\newcommand{\bX}{\mathbf{X}}
\newcommand{\bY}{\mathbf{Y}}

\newcommand{\bv}{\mathbf{v}}
\newcommand{\cS}{\ensuremath{\mathcal{S}}}

\newcommand{\cA}{\ensuremath{\mathcal{A}}}
\newcommand{\1}{\mathbbm{1}}
\newcommand{\I}{\mathbb{I}}

\newcommand{\cD}{\ensuremath{\mathcal{D}}}
\newtheorem{mydef}{Definition}

\usepackage{tikz}
\usetikzlibrary{arrows,backgrounds}
\usepgflibrary{shapes.multipart}

\title{
Optimizing generalization on the train set: a novel gradient-based framework to train parameters and hyperparameters simultaneously 
} 

\begin{document}

\maketitle 

\begin{abstract}
Generalization is a central problem in Machine Learning.
Most prediction methods require careful calibration of hyperparameters carried out on a hold-out \textit{validation} dataset to achieve generalization. The main goal of this paper is to present a novel approach based on a new measure of risk that allows us to develop novel fully automatic procedures for generalization. We illustrate the pertinence of this new framework in the regression problem. The main advantages of this new approach are: 
(i) it can simultaneously train the model and perform regularization in a single run of a gradient-based optimizer on all available data without any previous hyperparameter tuning; (ii) this framework can tackle several additional objectives simultaneously (correlation, sparsity,...) $via$ the introduction of regularization parameters. Noticeably, our approach transforms hyperparameter tuning as well as feature selection (a combinatorial discrete optimization problem) into a continuous optimization problem that is solvable via classical gradient-based methods
; (iii) the computational complexity of our methods is $O(npK)$ where $n,p,K$ denote respectively the number of observations, features and iterations of the gradient descent algorithm. We observe in our experiments a significantly smaller runtime for our methods as compared to benchmark methods for equivalent prediction score. Our procedures are implemented in PyTorch (code is available for replication).
\end{abstract}
\section*{Introduction and Related works}
Generalization is a central problem in machine learning. 
Regularized or constrained Empirical Risk Minimization (\textit{\texttt{ERM}}) is a popular approach to achieve generalization \cite{kukavcka2017regularization}. Ridge \cite{hoerl1970ridge}, \LAS \cite{tibshirani1996regression} and Elastic-net \cite{zou2005regularization} belong to this category.
The regularization term or the constraint is added in order to achieve generalization and to enforce some specific structures on the constructed model (sparsity, low-rank, coefficient positiveness,...). This usually involves introducing hyperparameters that need to be properly calibrated, which requires a good estimate of the generalization risk. 
The most common approach is data-splitting that partitions the available data into a {\it training/validation}-set. The {\it validation}-set is used to evaluate the generalization error of a model built using only the {\it training}-set. Another approach is based on unbiased estimation of the generalization error of a model (SURE \cite{stein1981estimation}, $AIC$ \cite{akaike1974new}, $C_p$-Mallows \cite{mallows2000some}) on the $training$-set.
Several hyperparameter tuning strategies are designed to achieve generalization: Grid-search, Random search \cite{bergstra2012random} or more advanced hyperparameter optimization techniques \cite{bergstra2011algorithms, bengio2000gradient,schmidhuber1987evolutionary}. For instance, BlackBox optimization \cite{brochu2010tutorial} is used when the evaluation function is not available \cite{lacoste2014sequential}. It includes in particular Bayesian hyperparametric optimization such as Thompson sampling \cite{movckus1975bayesian,snoek2012practical,thompson1933likelihood}.
Note that this technique either scales exponentially with the number of hyperparameters, or requires a smooth convex optimization space \cite{shahriari2015taking}. Highly non-convex optimization problems can be tackled by Population based methods (Genetic Algorithms\cite{chen2018autostacker,real2017large,olson2016automating}, Particle Swarm~\cite{lorenzo2017particle,lin2008particle}) for a high computational cost. Another family of advanced methods, called gradient-based techniques, take advantage of gradient optimization techniques \cite{domke2012generic} like our method. They fall into two categories, Gradient Iteration and Gradient approximation. Gradient Iteration directly computes the gradient w.r.t. hyperparameters on the training/evaluating graph. 
It requires differentiating a potentially lengthy optimization process which is known to be a major bottleneck \cite{pedregosa2016hyperparameter}. Gradient approximation is used to circumvent this difficulty, through implicit differentiation \cite{larsen1996design,bertrand2020implicit}. All these advanced methods require data-splitting and the evaluation of the trained model on a hold-out $validation$-set unlike our approach.\\
We can cite other methods that improve generalization during the training phase without using a hold-out $validation$-set. For instance, Stochastic Gradient Descent and the related batch learning techniques \cite{bottou1998online} achieve generalization by splitting the training data into a large number of subsets and compute the Empirical Risk (ER) on a different subset at each step of the gradient descent. This strategy converges to a good estimation of the generalization risk provided a large number of observations is available. Note that this method and the availability of massive datasets played a crucial role in the success of Deep neural networks. Although it has been shown that batch size has a positive impact on generalization \cite{he2019control}, it cannot maximize generalization on its own.
Model aggregation is another popular approach to achieve generalization. It concerns for instance Random Forest \cite{ho1995random,breiman2001random}, MARS \cite{friedman1991multivariate} and Boosting \cite{freund1995desicion}. This approach aggregates weak learners previously built using bootstrapped subsets of the $training$-set. The training time of these models is considerably lengthened when a large number of weak learners is considered, which is a requirement for improved generalization. Note that XGBOOST \cite{chen2016xgboost} combines a version of batch learning and model aggregation to train weak learners.\\
MARS, Random Forest, XGBOOST and Deep learning have obtained excellent results in Kaggle competitions and other machine learning benchmarks \cite{fernandez2014we, escalera2018neurips}.
However these methods still require regularization and/or constraints in order to generalize. It implies the introduction of numerous hyperparameters that need to be calibrated on an hold-out $validation$-set for instance \textit{via} Grid-search. Tuning these hyperparameters requires expensive human expertise
and/or computational resources.
We present a novel approach based on a different understanding of the generalization notion. This allows us to derive several training procedures that are fully automatic and do not require data splitting to achieve generalization. The underlying intuition is that we no longer see generalization as the ability of a model to perform well on unseen data, but rather as the ability to avoid finding pattern when none exist, we will refer to this situation as the \textit{non-informative} case.
The rest of the paper is organized as follows. In Section \ref{Sec:GER} we present our novel approach in the regression setting. Based on this approach, we develop in Section \ref{Sec:BKK} a framework for the linear regression setting and build procedures able to capture specific structures. In Section \ref{sec:Imp}, we carry out an extensive numerical study that highlights several interesting points about this new framework. Finally we discuss possible directions for future work in Section \ref{sec:con}.
\section{A novel approach for generalization}
\label{Sec:GER}
We place ourselves in the regression setting where our goal is to recover $\mathbb{E}[Y|\x]$ from a sample of observations of $(\x,Y)$. Let consider a {\it train}-sample $\cD= \left\lbrace (\x_1,Y_1),\cdots,(\x_n,Y_n) \right \rbrace$ of size $n$, where the pairs $(\x_i,Y_i)$ are independent and take values in $\mathcal{X}\times \mathcal{Y}$.
Let denote by $\mathbf{Y}$ the $n$-dimensional vector $\mathbf{Y}=(Y_i,\cdots,Y_n)^\top$ and by $\mathbf{X}$ the $n\times p$ \textit{design matrix} such that $\mathbf{X}^\top=(\x_1,\cdots,\x_n)$. In the regression setting, we often want to recover $\mathbb{E}[Y|\x]=f_\beta(X)$ from $\cD=(\mathbf{X},\mathbf{Y})$, where $f_\beta \,:\, \mathcal{X}\rightarrow \mathcal{Y}$ is a function which belongs to a considered class of models $\mathcal{F}$. This is usually done by minimizing the following empirical risk (\ER) w.r.t. the parameter $\beta$: $\ER(\mathbf{X},\mathbf{Y},\beta)=\frac{1}{n}\sum_{i=1}^n L(Y_i,f_{\beta}(\x_i))$, where $L\,:\, \mathcal{Y}\times \mathcal{Y}\rightarrow \mathbb{R}^{+}$ is a loss function.
In practice, the correlation between $\x$ and $Y$ is unknown and may actually be very weak. In that case, $\x$ provides very little information about $Y$ and we expect from a good procedure to avoid building a spurious connection between $\x$ and $Y$. Therefore, by understanding generalization as \say{\textit{do not fit the data in non-informative cases}}, we suggest creating a artificial dataset that preserves the marginal distributions, but such that the link between $\x$ and $Y$ has been completely removed. A simple way to do so is to construct an artificial one $\widetilde{\cD}=(\widetilde{\mathbf{X}},\widetilde{\mathbf{Y}}) = (\bX,\pi(\mathbf{Y}))$ by applying a permutation $\pi \in \mathfrak{S}_n$ (the set of permutations of $n$ points) on the components of $\bY$ of the initial dataset $\cD$ where for any $\mathbf{y}\in \mathbb{R}^n$, we set $\pi(\mathbf{y}) = (y_{\pi(1)},\ldots,y_{\pi(n)})^\top$. \\
Formally, it will result in the addition of a term to \ER. As the new dataset $\widetilde{\cD}$ is built directly on the $train$-set, generalization can be achieved without any data splitting. Formally, this amounts to adding to \ER\, a term that achieves generalization. We called this new risk measure the \GER. With this approach, novel criteria that need no hold-out $validation$-set can be developed and used in the same way as SURE, AIC, $C_p$-mallows to perform various tasks (enforce generalization, model selection, feature engineering). The \GER\, approach can also be used to derive novel procedures if we have access to a family of closed-form estimators $\{\widehat\beta_\theta\}_{\theta}$ depending on a regularization parameter $\theta$. From now on and for the sake of notation simplicity, we will write $\ER(\bX,\bY,\theta)$ for $ \ER(\bX,\bY,\widehat\beta_\theta)$.
\medskip
\begin{mdframed}
\underline{\textbf{\textit{\texttt{ER-G}}-APPROACH}}
\vspace{0.1cm}

Fix $T \in \mathbb{N}^*$. Let $\{\pi^t\}_{t=1}^T$ be $T$ permutations in $\mathfrak{S}_n$. 
(i) The \textbf{\GER\, criterion} is defined as
\begin{align}
\label{GERdef2}
\GER (\bX,\bY,\theta)&=\ER(\bX,\bY,\theta)+\frac{1}{T}\sum_{t=1}^T \left\vert \ER(\mathbf{X}_{0},\bY,\theta) - \ER(\bX,\pi^t(\bY),\theta)  \right\vert,
\end{align}
where $\mathbf{X}_{0} = (\1_n|\mathbf{0}_{n\times (p-1)})$ and $\1_n$ is the $one$ $n$-dimensional vector.

(ii) The \textbf{\GER\, procedure} is given for a family of closed-form solution $\{\widehat\beta(\theta)\}_{\theta}$ by 
\begin{eqnarray}\label{eq:pBKK1}
\begin{array}{lcl}
  \widehat \be^{\GER } =\widehat\be(\widehat\theta)&\text{with} 
  &\widehat{\theta} = \arg \min_{\theta} \GER (\bX,\bY,\theta).
\end{array}
\end{eqnarray}
\end{mdframed}
\paragraph{Discussion.}
The \GER\, criterion (\ref{GERdef2}) performs a trade-off between the first term which fits the data while the second term prevents overfitting.
The quantity $\ER(\mathbf{X}_{0},\bY,\theta)$ corresponds to the risk of the best estimator when no features $\x$ is included in the model.  For instance, for the regression linear model with the quadratic loss, this term is simply the standard deviation of $\bY$. Thus, the second term in (\ref{GERdef2}) states that a model should not perform better than the intercept model in non-informative cases.\\
An empirical study (See Figure \ref{fig:syntheticT}) about the impact on the generalization performances of the value of $T$ was carried out in the linear regression model considered in Section~\ref{Sec:BKK}. It revealed that this parameter has virtually no impact. Therefore, $T$ is not an hyperparameter that requires tuning. We set
 $T=30$ in our experiments 
which is largely sufficient to achieve generalization.
In addition, this approach allows us to enforce several \textbf{additional structures} simultaneously (sparsity, correlation, group sparsity, low-rank,...) just by considering an appropriate family of models encoded into $\theta$. We call $\theta$ the regularization parameter as we do not calibrate it on a hold-out $validation$-set which is a requirement for hyperparameters. Indeed, in our approach, we can tune $\theta$ directly on the $train$ set. 
Feature selection is by
essence a discrete optimization problem whereas the \GER\, criterion transforms this task into a continuous optimization problem (see Section~\ref{Sec:BKK}) that is solvable via an unique classical gradient-based methods. \\
A joint effort by the communities of electronic engineering, computer science and numerical optimization has resulted into the combined optimization of the software (graph-based computation, automatic differentiation) and hardware architecture (GPU's, TPU's) in order to compute gradients efficiently. Since our method is purely gradient based on a specifically designed derivation graph, it can immediately benefits from this optimized tensorized computation environment.
The message that we want to convey in this paper is
the following. It is possible to achieve generalization on the $train$-set $via$ the \GER\, criterion. Although we
illustrate the potential of the \ER-G\,approach only in the linear regression setting in Section \ref{Sec:BKK}, we believe it was interesting to provide here a general approach to investigate other Machine Leaning settings.
Of course, we are aware that adapting this approach to include other models will require significant work. Indeed,
the \GER\, procedures currently depends on a closed-form family of models.
We are currently developing an approach to remove this restriction and extend it to more general families of models and other settings such as the classification problem.
The \GER\, criterion (\ref{GERdef2}) contains two antagonistic terms which result into large oscillations on the loss landscape (w.r.t. $\theta$) in our experiments. This was due to an inappropriate choice of $L$ ($e.g.$ quadratic loss) that resulted in instability in the generalization performance. To remediate this issue, it is important to choose a uniformly continuous loss function so that the criterion does not oscillate too abruptly w.r.t. $\theta$. In that regard, the quadratic loss (not uniformly continuous) was not a good candidate and we used instead the square root of the quadratic loss in Section \ref{Sec:BKK}. Note that this loss is differentiable w.r.t. $\theta$ as long as $\bY\neq \bX \widehat{\beta}_{\theta}$ \cite{bunea2013group}. See the Appendix for other possible choices of loss functions. Moreover, we observed in our experiments that the loss function $L$ does not need to be convex to return interesting results (See Section \ref{Sec:BKK}). 
\section{Linear regression setting}
\label{Sec:BKK}
From now, we place ourselves in the linear regression setting where $\mathcal{X} = \mathbb{R}^p$, $\mathcal{Y}= \mathbb{R}$ and $\mathcal{F} = \left\lbrace f_{\beta}(\x) = \langle \x, \beta\rangle, \, \beta\in\mathbb{R}^p   \right\rbrace$ is the class of linear functions. We consider the linear regression model:
\begin{eqnarray}
\label{mod2}
 \bY = \bX \be^* + \bm{\e},
\end{eqnarray}
where $\bX^\top=(\x_1,\cdots,\x_n)$ is the $n\times p$ \textit{design matrix} and  the $n$-dimensional vectors $\bY=(Y_i,\cdots,Y_n)^\top$ and $\bm{\e}=(\e_1,\cdots,\e_n)^\top$ are respectively the response and the noise variables. Throughout this paper, the noise level $\sigma>0$ is unknown.
We specify the \GER\, approach in the linear model setting with the square root quadratic loss with $||\bv||_2=(\frac{1}{n}\sum_{i=1}^n v_i^2)^{1/2}$ for any $\bv=(v_1,\ldots,v_n)^\top\in \R^n$. Since our framework is defined for centered and rescaled response $\bY$, the standard deviation term $\ER(\mathbf{X}_{0},\bY,\theta)=1$.
\medskip
\begin{mdframed}
\underline{\textbf{\textit{\texttt{BKKs}}-FRAMEWORK}}\footnote{There is no particular meaning behind the name BKK as it was the result of a private joke.}
\vspace{0.1cm}
Fix $T \in \mathbb{N}^*$.  Let $\{\pi^t\}_{t=1}^T$ be $T$ permutations in $\mathfrak{S}_n$. 
(i) \textbf{\BKKs\, criterion} is defined as
\begin{small}\begin{align}
\label{eq:cBKKs}
\BKKs_\beta(\theta) &= \Vert \bY - \bX \be(\theta, \bX, \bY)\Vert_2 + \frac{1}{T} \sum_{t=1}^{T} \Bigg| 1 - \Vert \pi^t(\bY) - \bX\be(\theta, \bX, \pi^t(\bY)) \Vert_2  \Bigg|.
\end{align}
\end{small}

(ii) \textbf{\BKKs\, procedure} is given for a closed-form family of estimators $\{\be(\theta)\}_{\theta}$ by
\begin{small}\begin{eqnarray}
\label{eq:pBKKs}
\begin{array}{lcl}
  \widehat\be =\be(\widehat\theta, \bX, \bY)&\text{with} &\widehat \theta = \arg \min_{\theta}\BKKs_\beta(\theta).
\end{array}
\end{eqnarray}\end{small}
\end{mdframed}
\medskip
Minimizing the first term w.r.t. $\theta$ corresponds to the objective \say{\textit{Fit the data as well as possible}}. 
Minimizing the second term achieves the following: \say{\textit{Avoid overfitting by selecting only the intercept in \textit{non-informative cases}}}. 
Optimizing these two terms simultaneously on the $training$ set $\cD$ yields the generalization property of the \BKKs\, procedure.\\
From the adaptation of the \GER\, approach to the linear setting, we can derive several criteria to take into account different underlying structures. Ridge is one of the most popular closed-form family of estimators $\{\be^R(\lambda, \bX, \bY)\}_{\lambda>0}$:
\begin{align}
\label{Ridge}
\be^R(\lambda, \bX, \bY) = (\bX^\top \bX + \lambda \I_p)^{-1}\bX^\top \bY,\quad \lambda>0.
\end{align}
From a computational point of view, the matrix inversion is not expensive in our setting as long as the covariance matrix can fully fit on the GPU. Then, the extreme level of parallelization removes the dependency between computational time and the number of features as seen in our experimental results in the Appendix. Otherwise, additional learning schemes ($e.g$ feature bagging \cite{ho1998random}, block batch learning \cite{xu2015block}, Least Mean  Squares Solvers \cite{maalouf2019fast}) should be implemented.\\
From (\ref{Ridge}), we can derive our first criterion by taking $\BKK(\lambda):=\BKKs_{\beta^R}(\lambda)$. The \BKK\, procedure is designed to tackle strongly correlated features. The \BKKs\, criterion may be seen as a novel hyperparameter tuning criterion without hold-out $validation$-set. Experiments for the Ridge family \eqref{Ridge} reveal that our approach achieved significantly faster running times over cross-validation for equivalent generalization performances.

A novel family of closed-form estimators is designed to enforce sparsity in the trained model. To this end, we introduce a quasi-sparsifying operator in Definition \ref{sparsefunction}.

\begin{mydef}
\label{sparsefunction} 
 Let $\{\beta^{\cS}(\lambda,\kappa,\spa,\bX,\bY)\}_{(\lambda,\kappa,\spa)\in\R^*_+\times\R^*_+\times\R^p}\,\,$ be a family closed-form of estimators defined as follows:
\begin{align}
\label{Sbeta}
\be^{\cS}(\lambda,\kappa,\spa, \bX, \bY) = \cS(\kappa,\spa) \beta^R(\lambda, \bX \cS(\kappa,\spa), \bY),
\end{align}
where $\be^R$ is defined in \eqref{Ridge}, the \textbf{\textit{quasi-sparsifying}} function $\cS:\R^*_+ \times\R^p \rightarrow ]0,1[^{p\times p}$ is $s.t.$
$$\cS(\kappa,\spa)=\text{diag}\left(\cS_1(\kappa,\spa),\cdots,\cS_p(\kappa,\spa)\right),$$
with for any $j=1,\cdots,p$ \,\, , $\overline \spa=\frac{1}{p}\sum_{i=1}^p \spa_i$ and $\sigma_\gamma^2=\sum_{i=1}^p (\spa_i-\overline \spa)^2$

$$
\cS_j \ : \R^*_+ \times\R^p \rightarrow ]0,1[\; : \; 
(\kappa,\spa) \mapsto\cS_j (\kappa,\spa)=\left(1+e^{-\kappa \times (\sigma_\gamma^2 + 10^{-2})(\spa_j - \overline \spa)}\right)^{-1}.
$$
\end{mydef}

The new family \eqref{Sbeta} enforces sparsity on the regression vector but also directly onto the design matrix. Hence it can be seen as a combination of data-preprocessing (performing feature selection) and model training (using the ridge estimator).\\
Noticeably, the \say{\textit{quasi-sparsifying}} trick transforms feature selection (a discrete optimization problem) into a continuous optimization problem that is solvable via classical gradient-based methods.
The function $\cS$ produces diagonal matrices with diagonal coefficients in $]0,1[$. While the sigmoid function $\cS_j$ cannot take values 0 or 1, we note however that for very small or large values of $\gamma_j$, the value of the corresponding diagonal coefficient of $S(\kappa,\gamma)$ is extremely close to $0$ or $1$. In those cases, because of the finite \textit{in silico} variable precision, we observe in our numerical experiments that the sigmoid function essentially yields 0's or 1's. Hence, the resulting model $\beta^{\cS}$ is sparse. From (\ref{Sbeta}), we can derive our second criterion by taking $\cS\BKK(\lambda,\kappa,\gamma):=\BKKs_{\beta^{\cS}}(\lambda,\kappa,\gamma)$.

By aggregating the families \eqref{Ridge} and \eqref{Sbeta}, we can build another closed-form family of estimators (Definition~\ref{ABKK}). It essentially  consists in an interpolation between $\beta^R$ and $\beta^\cS$ estimators, where the level of interpolation is quantified via the introduction of a new regularization parameter $\mu \in \R$.
\begin{mydef}
\label{ABKK}
 Let $\{\beta^{\cA}(\lambda,\kappa,\spa,\mu, \bX, \bY)\}_{(\lambda,\kappa,\spa,\mu)\in\R^*_+\times\R^*_+\times\R^p\times\R}\,\,$ be a closed-form family of estimators defined as follows:
\begin{align}
\label{Abeta}
\be^{\cA}(\theta, \bX,\bY) =  \Sig(\Emu) \times \beta^R(\lambda, \bX,\bY) + (1 - \Sig(\Emu)) \times \be^{\cS}(\lambda, \kappa,\spa, \bX,\bY),
\end{align}
where $\be^R(\lambda,\bY)$ and $\be^{\cS}(\lambda,\kappa, \spa, \bX, \bY)$ are defined in \eqref{Ridge} and \eqref{Sbeta} respectively and $\Sig$ is the sigmoid\footnote{For $\Emu\in\R$,  $\Sig(\Emu)$ takes values in $(0,1)$ and is actually observed in practice to be close to $0$ or $1$.}.
\end{mydef}
From (\ref{Abeta}), we can derive our last criterion by taking $\cA\BKK(\lambda,\kappa,\gamma,\mu):=\BKKs_{\beta^{\cA}}(\lambda,\kappa,\gamma,\mu)$ which can handle both correlation and sparsity.  
\paragraph{Discussion.}
The \BKKs\, criteria enable us to develop fully automatic procedures to tune regularization parameters while simultaneously training the model in a single run of the gradient descent algorithm without a hold-out $validation$ set. The computational complexity of our methods is $O(n(p+r)K)$ where $n,p,r,K$ denote respectively the number of observations, features, regularization parameters and iterations of the gradient descent algorithm. Note that the computational complexity of our method grows only arithmetically w.r.t. the number of regularization parameters.
In this paper, we choose to compute the \BKKs\, procedures with ADAM. Although \cS\BKK\, and \cA\BKK\, are highly non-convex, we can still use ADAM without any significant increase in the number of iterations to achieve convergence \cite{kingma2014adam}. We observed in our experiments that at most a few dozen iterations are required to compute our procedures resulting in a significantly faster running time.
An even more striking fact is that the initial values (see Table~\ref{Tab:parameter} for the main ones) to implement our \BKKs\, procedures remain the same for all the datasets we considered for consistently good prediction performances. This is not usually the case with other approaches.
\begin{table}[http!]
\begin{center}\centering
\begin{tabular}{|l||l|}
\hline
\footnotesize{\textbf{Optimization parameters}}&  \footnotesize{\textbf{Parameter initialization}}\\
\hline
\hline
\begin{tabular}{l|l}
\footnotesize{\textbf{Tolerance}} &  $10^{-4}$\\
\footnotesize{\textbf{Max. iter.}} &  $10^{3}$\\
\footnotesize{\textbf{Learning rate}} &  $0.5$\\
\footnotesize{\textbf{Adam $\beta_1$}} &  $0.5$\\
\footnotesize{\textbf{Adam $\beta_2$}} &  $0.9$\\
\end{tabular} &
\begin{tabular}{l|l}
\footnotesize{\textbf{$T$}} & $30$ \\
\footnotesize{\textbf{$\lambda$}} & $10^{3}$ \\
\footnotesize{\textbf{$\spa$}} & $0_p$ \\
\footnotesize{\textbf{$\kappa$}} & $0.1$ \\
\footnotesize{\textbf{$\mu$}} & $0$ \\
\end{tabular} \\
\hline
\end{tabular}
\end{center}
\caption{Parameters for the \BKKs\, procedures.}
\label{Tab:parameter}
\end{table}
\section{Numerical experiments}\label{sec:Imp}
In this section, we compare our \BKKs\, procedures against cross-validation on Ridge, Lasso and Elastic-net (implemented as RidgeCV,
LassoCV and ElasticnetCV in Scikit-learn \cite{pedregosa2011scikit}) 
on simulated and real datasets. Our procedures are implemented in PyTorch (\cite{paszke2019pytorch}) on the centered and rescaled 
response $\bY$. Complete details and results can be found in the Appendix. Our Python code is released as an open source package for replication on \href{https://github.com/AnonymousSubmissionNeurips2020/Supplementary-Material.git}{github/AnonymousSubmissionNeurips2020/}. 
\paragraph{Synthetic data.}
We generate observations $(\x,Y) \in \R^p\times \R$, $p=80$ $s.t.$ 
$
Y = \x^\top \beta^* + \epsilon,
$
with $\epsilon\sim\mathcal N(0,\sigma)$, $\sigma=10$ or $50$. We consider three different scenarii. \textbf{Scenario A} (correlated features) corresponds to the case where the \LAS is prone to fail and Ridge should perform better. \textbf{Scenario B} (sparse setting) corresponds to a case known as favorable to \LAS. \textbf{Scenario C} combines sparsity and correlated features. 
For each scenario we sample a $train$-dataset of size $n_{train}=100$ and a $test$-dataset of size $n_{test}=1000$.\\
For each scenario, we perform $M=100$ repetitions of the data generation process to produce $M$ pairs of $train$/$test$ datasets. Details on the data generation process can be found in the Appendix.
\paragraph{Real data.}
We test our methods on several commonly used real datasets (UCI \cite{asuncion2007uci} and Svmlib \cite{chang2011libsvm} repositories). See Appendix for more details. Each selected UCI dataset is splitted randomly into a $80\%$ $train$-dataset and a $20\%$ $test$-dataset. We repeat this operation $M=100$ times to produce $M$ pairs of $train$/$test$ datasets.\\
In order to test our procedures in the setting $n \leq p$, we selected, from Svmlib, the news20 dataset which contains a $train$ and a $test$ dataset. We fixed the number of features $p$ and we sample six new 20news $train$-datasets  of different sizes $n$ from the initial news20 $train$-dataset. For each size $n$ of dataset, we perform $M=100$ repetitions of the sampling process to produce $M$ $train$ datasets. We kept the initial $test$-set for the evaluation of the generalization performances.

\paragraph{Performances evaluation criteria.}

For each procedure and each $train$ dataset, we construct the corresponding model $\widehat{\be}_{train}$ based on $\mathcal D_{train}=(\bX_{train},\bY_{train})$. The generalization performances of all the considered procedures are evaluated using the hold-out $test$-sets $\mathcal D_{test}=(\bX_{test},\bY_{test})$ by computing their $\mathbf{R^2}$-scores as follows
\begin{eqnarray}
\label{R2}
\mathbf{R^2}(\widehat{\be}_{train})=1- \frac{\Vert \bY_{test} - \bX_{test}\,\widehat{\be}_{train} \Vert_2}{\Vert \bY_{test} -\overline{\bY}_{test}\1_n \Vert_2}\,(\leq 1),
\end{eqnarray}
where $\overline{\bY}_{test}$ is the empirical mean of $\bY_{test}$. 

The higher the value of the $\mathbf{R^2}$-score, the better the generalization performance of a procedure.

For the $M=100$ repetitions of each synthetic scenario and each real dataset, we implement all the procedures and record their $\mathbf{R^2}$-scores and running times.\\
Boxplots summarize our findings. The empirical mean is represented with a green triangle.
To check for statistically significant margin in $\mathbf{R^2}$-scores and running times between different procedures, we use the Mann-Whitney test (as detailed in \cite{kruskal1957historical} and implemented in scipy \cite{virtanen2020scipy}). The boxplots highlighted in yellow correspond to the best procedures according to the Mann-Whitney (MW) test.

\paragraph{Comparison of the different procedures.}

\begin{figure}[http!]
\centering
\includegraphics[scale=0.5]{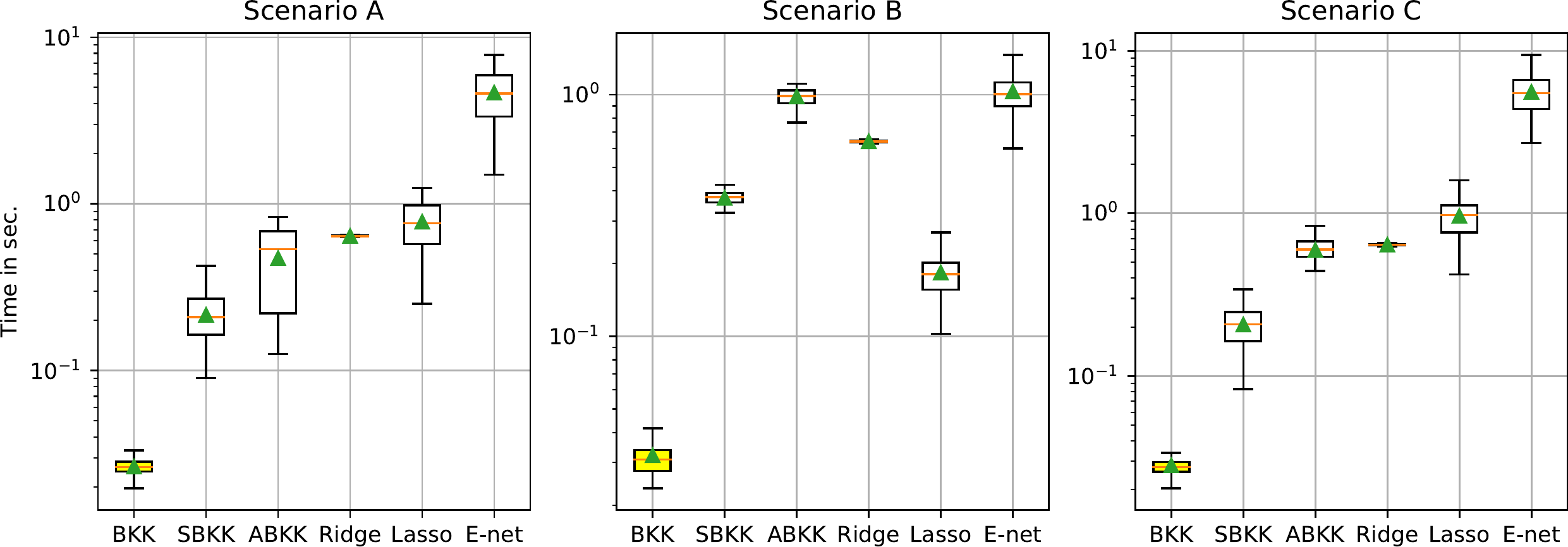}
\caption{\textbf{Synthetic data:} running times in seconds.}
\label{fig:syntheticTime}
 \end{figure}

\begin{figure}[http]
\centering
\includegraphics[scale=0.45]{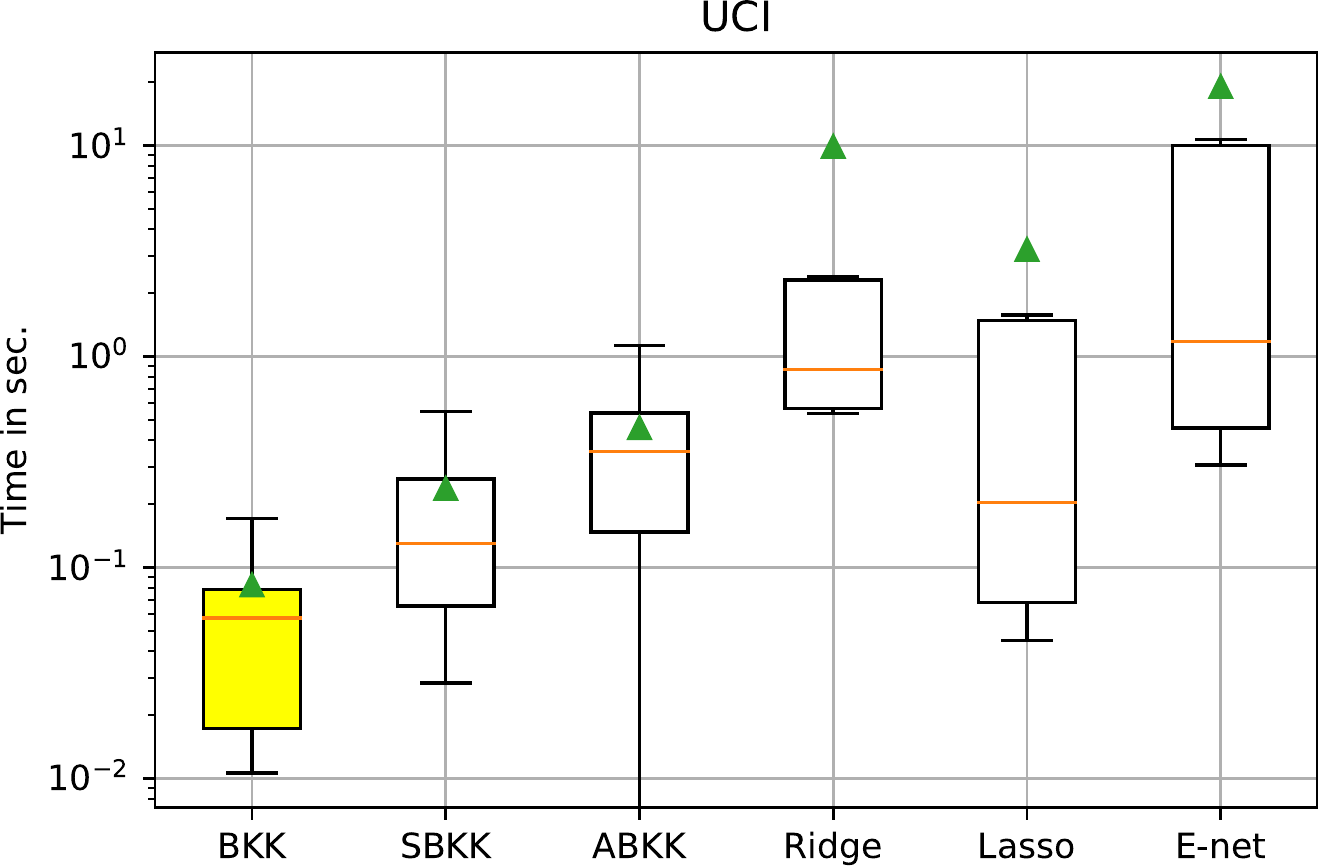}
\includegraphics[scale=0.45]{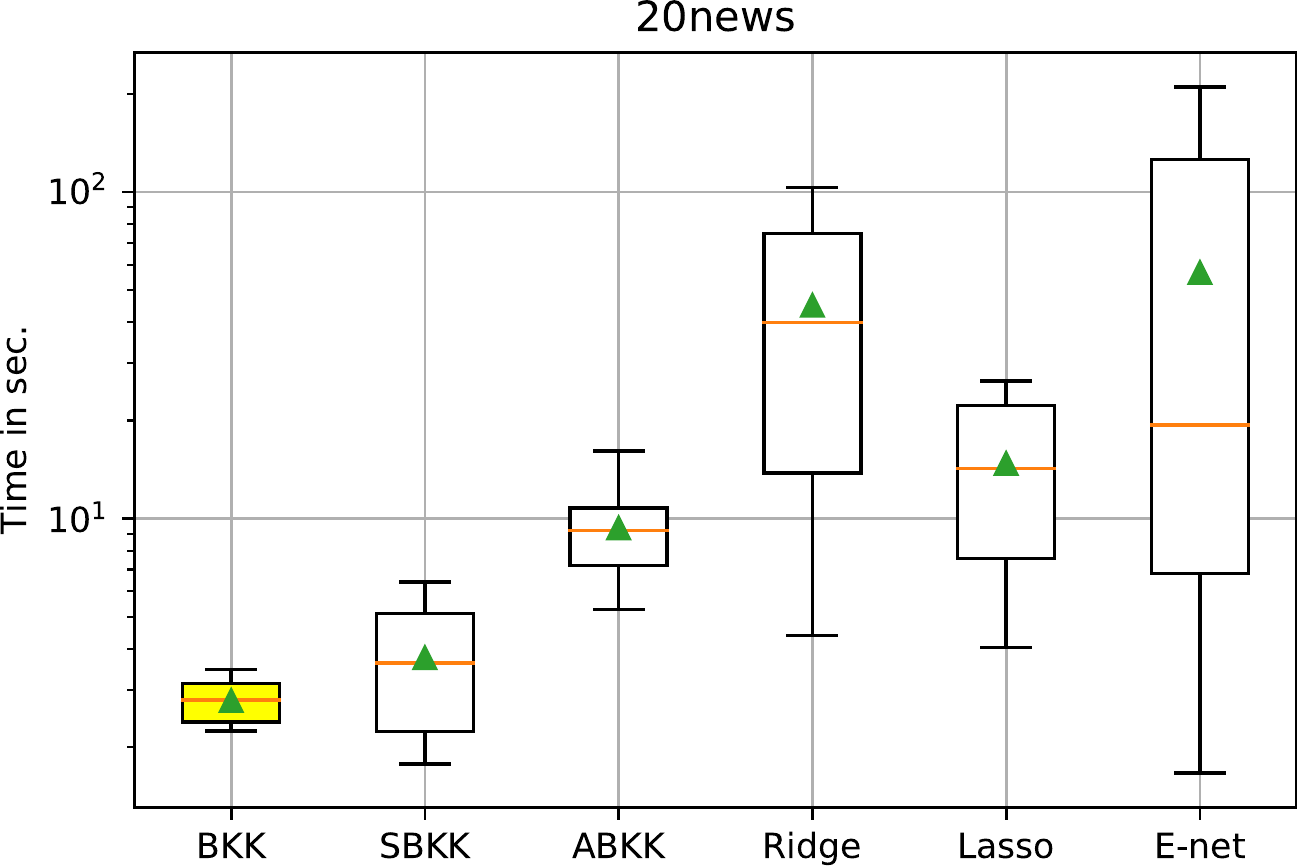}
\caption{\textbf{UCI data} (left) and \textbf{20news data} (right): running times in seconds.}
\label{fig:graphtime}
\end{figure}

Figure~\ref{fig:syntheticTime}  and \ref{fig:graphtime} concerns the running times of each procedure respectively on the synthetic datasets and on the real datasets. \\
For the synthetic data, the running times of the \BKKs\, procedures are always significantly smaller than those of the cross-validated benchmark procedures for equivalent to better $\mathbf{R^2}$-scores (see Appendix for $\mathbf{R^2}$-scores results). \\ 
 For the UCI datasets, the \BKKs\, always obtain the highest $\mathbf{R^2}$-scores according to the MW test. For the 20news datasets, the \BKKs\, are always within $0.05$ of the best. Figure \ref{fig:graphtime} reveals that the \BKKs\, are always significantly faster to compute for all real datasets according to the MW test. The \BKK\, procedures achieved an average speed-up up to $20$ times over the benchmark procedures  for the 20news datasets and up to $123$ times for the UCI datasets.

\begin{figure}[http!]
\centering
\includegraphics[scale=0.45]{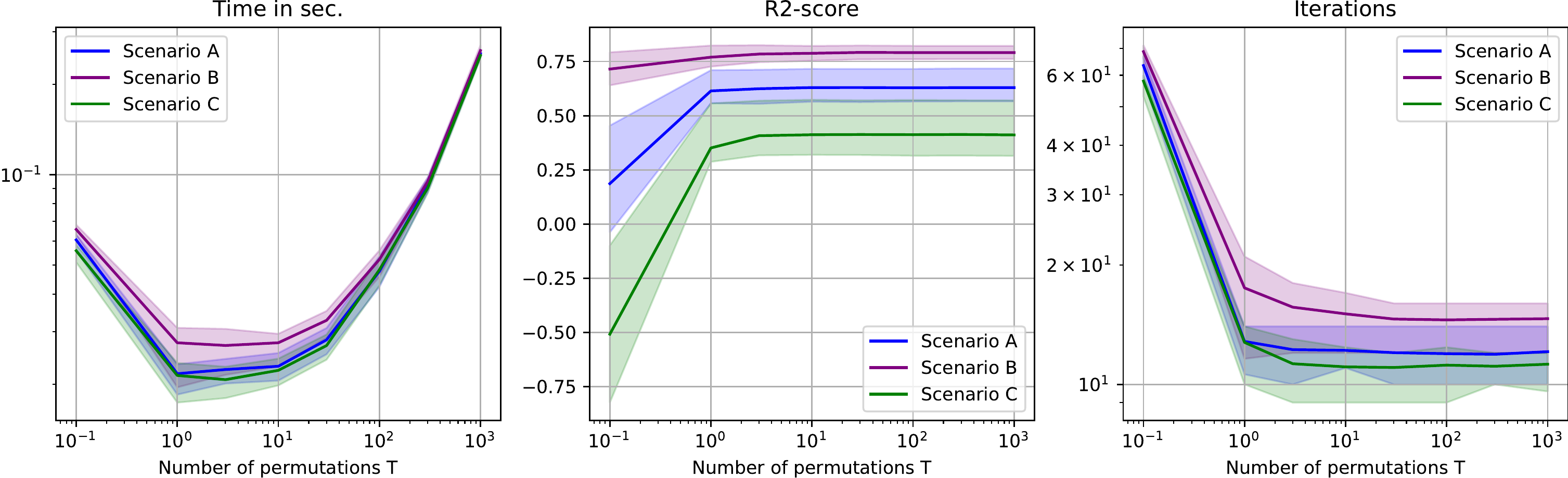}
\caption{\textbf{Synthetic data:} impact of $T$ on the \BKKs\, procedures. } 
\label{fig:syntheticT}
\end{figure}

\paragraph{Impact of parameter $T$ and number of iterations.}
Figure \ref{fig:syntheticT} plots the impact of parameter $T$ on the performances of all the \BKKs\, procedures on the synthetic data.

We observe that the generalization performance ($\mathbf{R^2}$-score) increases significantly as soon as $T=1$. Starting from $T\approx 10$, the $\mathbf{R^2}$-score has converged to its maximum value. An even more striking phenomenon is the gain observed in the running time when we add $T$ permutations (for $T$ in the range from $1$ to approximately $100$) as compared to the usual risk $\ER$ ($T=0$). Note that larger values of $T$ are neither judicious nor needed in this approach.
It is also a pleasant surprise that the needed number of iterations for ADAM to converge is divided by $3$ starting from the first added permutation ($T=1$). Furthermore, the number of iterations remained stable (below 20) starting from $T=1$. 

\begin{figure}[htp]
\centering
\includegraphics[scale=0.35]{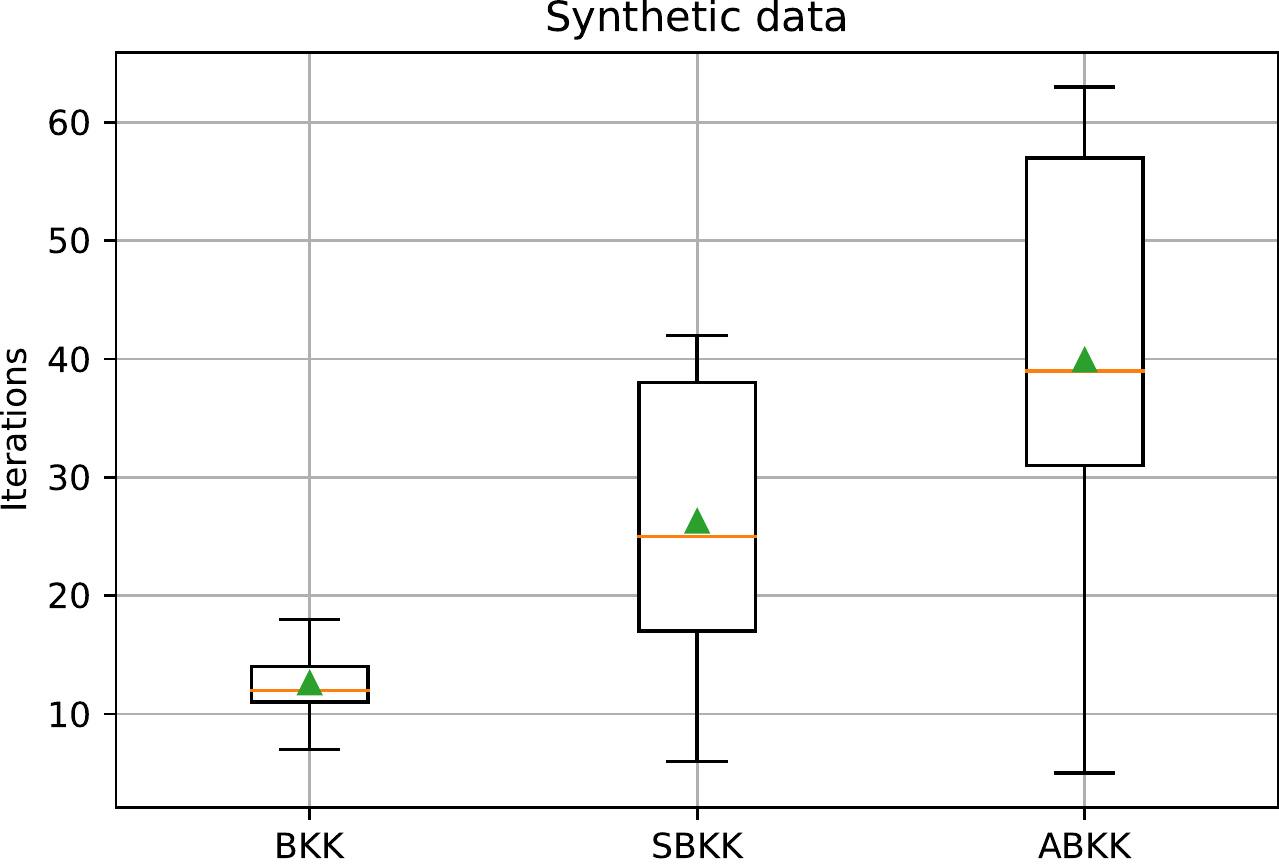}
\includegraphics[scale=0.35]{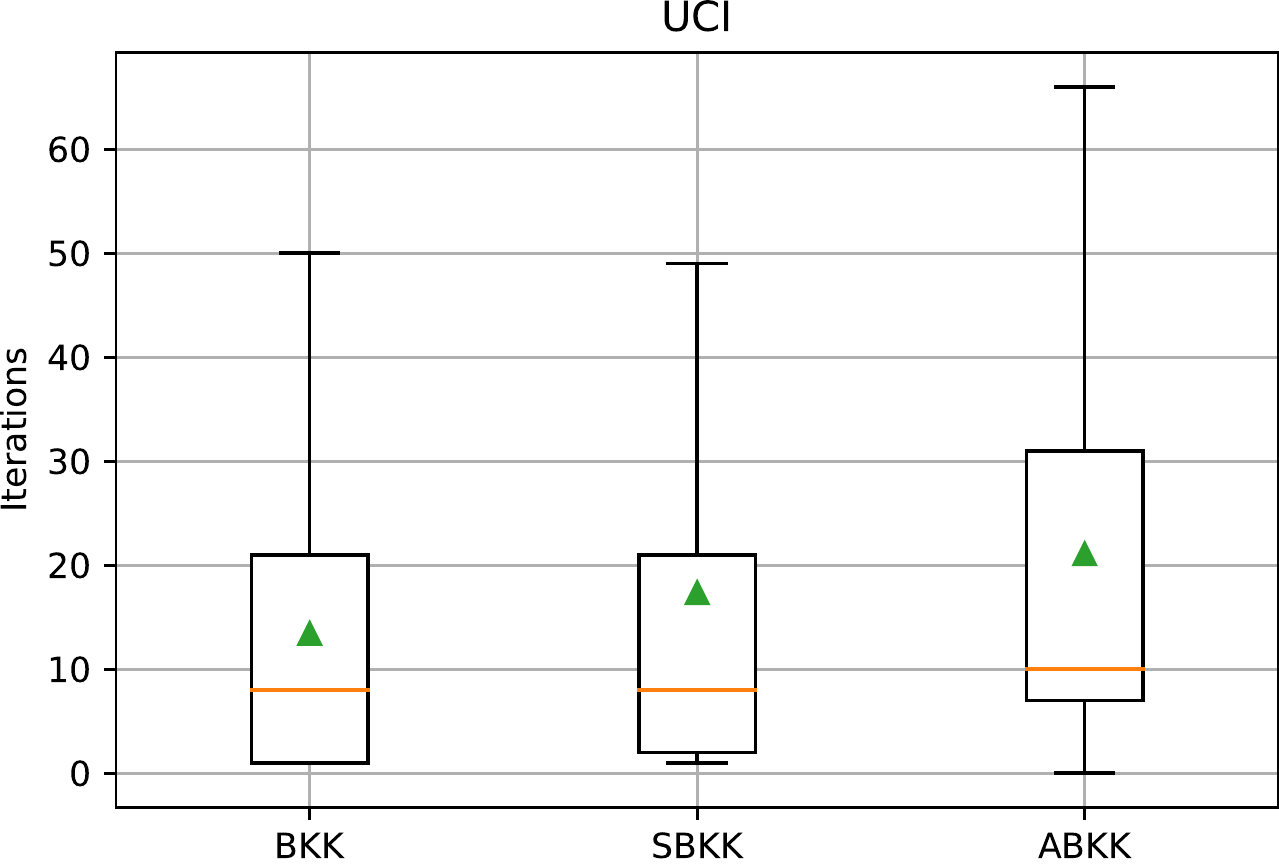}
\includegraphics[scale=0.35]{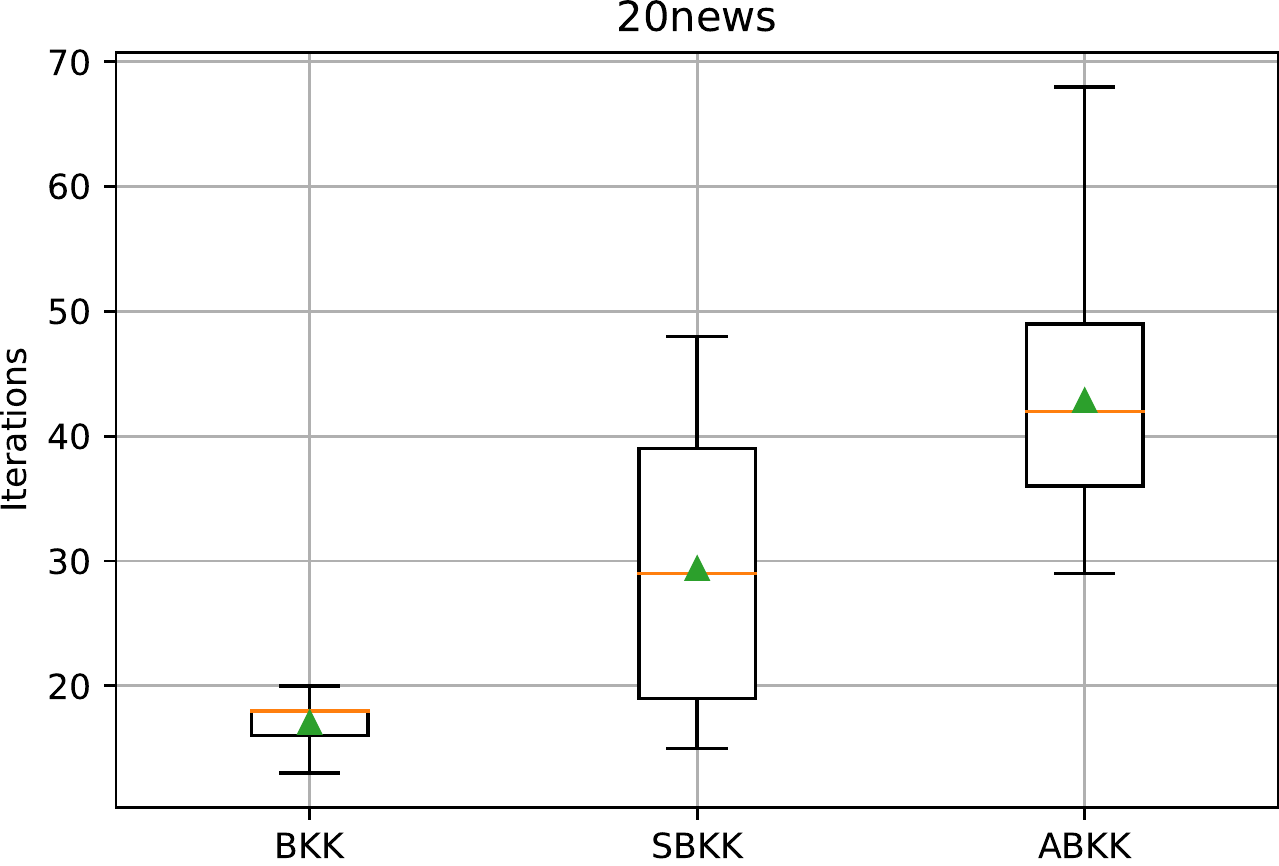}
\caption{\textbf{Synthetic, UCI and 20news data}:  Number of iterations}
\label{fig:itergraph}
\end{figure}

We fixed $T=30$ in our experiments even though $T=10$ may have been sufficient. Figure \ref{fig:itergraph} contains the boxplots of the number of ADAM iterations for the \BKKs\, procedures on the real datasets.
The number of iterations required for convergence is always about a few dozen in our experiments even for \cS\BKK\, and \cA\BKK\, which are highly non-convex. This was already observed in other non-convex settings \cite{kingma2014adam}. 

\section{Conclusion and future work}
\label{sec:con}
In this paper, we introduce in the regression setting the new \GER\, approach based on a different understanding of generalization. Exploiting this idea, criteria and procedures can be derived and implemented with a low computational complexity, $via$ gradient descent applied only once on the $train$-set without any hold-out $validation$-set. Within \GER, additional structures can taken into account without any significant increase in the computational complexity.\\
By applying \GER\, in the specific linear model setting, we developed a new framework in which several new criteria and procedures were derived. The resulting \BKKs\, are compatible with gradient-based optimization methods which fully exploit automatic graph-differentiation libraries (such as pytorch \cite{paszke2017automatic} and tensorflow \cite{tensorflow}). Noticeably, we ran these experiments using ADAM with a set parameters and stopping criterion that are outside of the range of values commonly found in the literature (far from PyTorch default parameters). Training new models in the \BKKs\, framework requires no tedious initialization. Indeed, we used the same fixed hyperparameters for ADAM and initialization values of the regularization parameters (see Table \ref{Tab:parameter}) for all the datasets we considered. Empirical investigations on the \BKKs\, procedures revealed substantial gains in running time while matching the generalization performances of cross-validated benchmark procedures. Moreover, ADAM converges after a very small number of iterations even for non-convex criteria (\cS\BKK\, and \cA\BKK).\\
The values of Adam parameters and the striking gain in running time due to the introduction of permutations, especially their impact on the behavior of the gradient descent methods, may deserve further investigation. Other common optimizers (ADAdelta\cite{dauphin2015equilibrated}, ADAboost \cite{freund1995desicion}, RMSProp introduced in \cite{graves2013generating}) will also be considered as well as techniques commonly used in deep learning (Stochastic Gradient Descent\cite{bottou1998online}, batch learning \cite{he2019control}, cycling learning rates \cite{smith2017cyclical}). Preliminary results suggest that significant gains in generalization performances should be added to the advantages already highlighted in this paper. In ongoing work, we also adapt the \GER\, approach to models with no closed-form estimators and classification tasks. 
In future work, we will investigate extensions of this method to non-linear models and times series. We believe our approach can also be extended to other types of data (image, text, graphs).
Note that our procedures give a promising direction to design fast differentiable optimization methods for the regularization of neural networks as our computational complexity is arithmetic w.r.t. the number of hyperparameters. Deep neural networks  produce state-of-the-art results on most machine learning benchmarks \cite{escalera2018neurips,he2016deep,klambauer2017self,krizhevsky2012imagenet,silver2016mastering,simonyan2014very,szegedy2015going}, but 
it is not yet the case on the U.C.I benchmark (as described in \cite{fernandez2014we}) for small datasets (less than 1000 observations as shown in \cite{klambauer2017self}). Moreover, training Deep neural networks requires heavy computational resources and manual supervision by experts given the huge number of hyperparameters to tune when working on new problems (see \cite{smith2018disciplined}). Our \GER\, approach may potentially lead to significant improvements in generalization performance of Deep neural networks and help accelerate their democratization beyond circles of experts.

\newpage

\bibliographystyle{plain}
\bibliography{BiblioPapier1}

\begin{thebibliography}{10}

\bibitem{tensorflow}
Martin Abadi, Ashish Agarwal, Paul Barham, Eugene Brevdo, Zhifeng Chen, Craig
  Citro, Greg~S. Corrado, Andy Davis, Jeffrey Dean, Matthieu Devin, Sanjay
  Ghemawat, Ian Goodfellow, Andrew Harp, Geoffrey Irving, Michael Isard,
  Yangqing Jia, Rafal Jozefowicz, Lukasz Kaiser, Manjunath Kudlur, Josh
  Levenberg, Dandelion Mane, Rajat Monga, Sherry Moore, Derek Murray, Chris
  Olah, Mike Schuster, Jonathon Shlens, Benoit Steiner, Ilya Sutskever, Kunal
  Talwar, Paul Tucker, Vincent Vanhoucke, Vijay Vasudevan, Fernanda Viegas,
  Oriol Vinyals, Pete Warden, Martin Wattenberg, Martin Wicke, Yuan Yu, and
  Xiaoqiang Zheng.
\newblock {TensorFlow}: Large-scale machine learning on heterogeneous systems,
  2015.
\newblock Software available from tensorflow.org.

\bibitem{akaike1974new}
Hirotugu Akaike.
\newblock A new look at the statistical model identification.
\newblock {\em IEEE transactions on automatic control}, 19(6):716--723, 1974.

\bibitem{asuncion2007uci}
Arthur Asuncion and David Newman.
\newblock Uci machine learning repository, 2007.

\bibitem{bengio2000gradient}
Yoshua Bengio.
\newblock Gradient-based optimization of hyperparameters.
\newblock {\em Neural computation}, 12(8):1889--1900, 2000.

\bibitem{bergstra2012random}
James Bergstra and Yoshua Bengio.
\newblock Random search for hyper-parameter optimization.
\newblock {\em Journal of machine learning research}, 13(Feb):281--305, 2012.

\bibitem{bergstra2011algorithms}
James~S Bergstra, R{\'e}mi Bardenet, Yoshua Bengio, and Bal{\'a}zs K{\'e}gl.
\newblock Algorithms for hyper-parameter optimization.
\newblock In {\em Advances in neural information processing systems}, pages
  2546--2554, 2011.

\bibitem{bertrand2020implicit}
Quentin Bertrand, Quentin Klopfenstein, Mathieu Blondel, Samuel Vaiter,
  Alexandre Gramfort, and Joseph Salmon.
\newblock Implicit differentiation of lasso-type models for hyperparameter
  optimization.
\newblock {\em arXiv preprint arXiv:2002.08943}, 2020.

\bibitem{bottou1998online}
L{\'e}on Bottou.
\newblock Online learning and stochastic approximations.
\newblock {\em On-line learning in neural networks}, 17(9):142, 1998.

\bibitem{breiman2001random}
Leo Breiman.
\newblock Random forests.
\newblock {\em Machine learning}, 45(1):5--32, 2001.

\bibitem{brochu2010tutorial}
Eric Brochu, Vlad~M Cora, and Nando De~Freitas.
\newblock A tutorial on bayesian optimization of expensive cost functions, with
  application to active user modeling and hierarchical reinforcement learning.
\newblock {\em arXiv preprint arXiv:1012.2599}, 2010.

\bibitem{bunea2013group}
Florentina Bunea, Johannes Lederer, and Yiyuan She.
\newblock The group square-root lasso: Theoretical properties and fast
  algorithms.
\newblock {\em IEEE Transactions on Information Theory}, 60(2):1313--1325,
  2013.

\bibitem{chang2011libsvm}
Chih-Chung Chang and Chih-Jen Lin.
\newblock Libsvm: A library for support vector machines.
\newblock {\em ACM transactions on intelligent systems and technology (TIST)},
  2(3):1--27, 2011.

\bibitem{chen2018autostacker}
Boyuan Chen, Harvey Wu, Warren Mo, Ishanu Chattopadhyay, and Hod Lipson.
\newblock Autostacker: A compositional evolutionary learning system.
\newblock In {\em Proceedings of the Genetic and Evolutionary Computation
  Conference}, pages 402--409, 2018.

\bibitem{chen2016xgboost}
Tianqi Chen and Carlos Guestrin.
\newblock Xgboost: A scalable tree boosting system.
\newblock In {\em Proceedings of the 22nd acm sigkdd international conference
  on knowledge discovery and data mining}, pages 785--794, 2016.

\bibitem{dauphin2015equilibrated}
Yann Dauphin, Harm De~Vries, and Yoshua Bengio.
\newblock Equilibrated adaptive learning rates for non-convex optimization.
\newblock In {\em Advances in neural information processing systems}, pages
  1504--1512, 2015.

\bibitem{domke2012generic}
Justin Domke.
\newblock Generic methods for optimization-based modeling.
\newblock In {\em Artificial Intelligence and Statistics}, pages 318--326,
  2012.

\bibitem{escalera2018neurips}
Sergio Escalera and Ralf Herbrich.
\newblock The neurips’18 competition.

\bibitem{fernandez2014we}
Manuel Fern{\'a}ndez-Delgado, Eva Cernadas, Sen{\'e}n Barro, and Dinani Amorim.
\newblock Do we need hundreds of classifiers to solve real world classification
  problems?
\newblock {\em The journal of machine learning research}, 15(1):3133--3181,
  2014.

\bibitem{freund1995desicion}
Yoav Freund and Robert~E Schapire.
\newblock A desicion-theoretic generalization of on-line learning and an
  application to boosting.
\newblock In {\em European conference on computational learning theory}, pages
  23--37. Springer, 1995.

\bibitem{friedman1991multivariate}
Jerome~H Friedman.
\newblock Multivariate adaptive regression splines.
\newblock {\em The annals of statistics}, pages 1--67, 1991.

\bibitem{graves2013generating}
Alex Graves.
\newblock Generating sequences with recurrent neural networks.
\newblock {\em arXiv preprint arXiv:1308.0850}, 2013.

\bibitem{he2019control}
Fengxiang He, Tongliang Liu, and Dacheng Tao.
\newblock Control batch size and learning rate to generalize well: Theoretical
  and empirical evidence.
\newblock In {\em Advances in Neural Information Processing Systems}, pages
  1141--1150, 2019.

\bibitem{he2016deep}
Kaiming He, Xiangyu Zhang, Shaoqing Ren, and Jian Sun.
\newblock Deep residual learning for image recognition.
\newblock In {\em Proceedings of the IEEE conference on computer vision and
  pattern recognition}, pages 770--778, 2016.

\bibitem{ho1995random}
Tin~Kam Ho.
\newblock Random decision forests.
\newblock In {\em Proceedings of 3rd international conference on document
  analysis and recognition}, volume~1, pages 278--282. IEEE, 1995.

\bibitem{ho1998random}
Tin~Kam Ho.
\newblock The random subspace method for constructing decision forests.
\newblock {\em IEEE transactions on pattern analysis and machine intelligence},
  20(8):832--844, 1998.

\bibitem{hoerl1970ridge}
Arthur~E Hoerl and Robert~W Kennard.
\newblock Ridge regression: Biased estimation for nonorthogonal problems.
\newblock {\em Technometrics}, 12(1):55--67, 1970.

\bibitem{kingma2014adam}
Diederik~P Kingma and Jimmy Ba.
\newblock Adam (2014), a method for stochastic optimization.
\newblock In {\em Proceedings of the 3rd International Conference on Learning
  Representations (ICLR), arXiv preprint arXiv}, volume 1412, 2014.

\bibitem{klambauer2017self}
G{\"u}nter Klambauer, Thomas Unterthiner, Andreas Mayr, and Sepp Hochreiter.
\newblock Self-normalizing neural networks.
\newblock In {\em Advances in neural information processing systems}, pages
  971--980, 2017.

\bibitem{krizhevsky2012imagenet}
Alex Krizhevsky, Ilya Sutskever, and Geoffrey~E Hinton.
\newblock Imagenet classification with deep convolutional neural networks.
\newblock In {\em Advances in neural information processing systems}, pages
  1097--1105, 2012.

\bibitem{kruskal1957historical}
William~H Kruskal.
\newblock Historical notes on the wilcoxon unpaired two-sample test.
\newblock {\em Journal of the American Statistical Association},
  52(279):356--360, 1957.

\bibitem{kukavcka2017regularization}
Jan Kuka{\v{c}}ka, Vladimir Golkov, and Daniel Cremers.
\newblock Regularization for deep learning: A taxonomy.
\newblock {\em arXiv preprint arXiv:1710.10686}, 2017.

\bibitem{lacoste2014sequential}
Alexandre Lacoste, Hugo Larochelle, Mario Marchand, and Fran{\c{c}}ois
  Laviolette.
\newblock Sequential model-based ensemble optimization.
\newblock In {\em Proceedings of the Thirtieth Conference on Uncertainty in
  Artificial Intelligence}, pages 440--448, 2014.

\bibitem{larsen1996design}
Jan Larsen, Lars~Kai Hansen, Claus Svarer, and M~Ohlsson.
\newblock Design and regularization of neural networks: the optimal use of a
  validation set.
\newblock In {\em Neural Networks for Signal Processing VI. Proceedings of the
  1996 IEEE Signal Processing Society Workshop}, pages 62--71. IEEE, 1996.

\bibitem{lin2008particle}
Shih-Wei Lin, Kuo-Ching Ying, Shih-Chieh Chen, and Zne-Jung Lee.
\newblock Particle swarm optimization for parameter determination and feature
  selection of support vector machines.
\newblock {\em Expert systems with applications}, 35(4):1817--1824, 2008.

\bibitem{lorenzo2017particle}
Pablo~Ribalta Lorenzo, Jakub Nalepa, Michal Kawulok, Luciano~Sanchez Ramos, and
  Jos{\'e}~Ranilla Pastor.
\newblock Particle swarm optimization for hyper-parameter selection in deep
  neural networks.
\newblock In {\em Proceedings of the genetic and evolutionary computation
  conference}, pages 481--488, 2017.

\bibitem{maalouf2019fast}
Alaa Maalouf, Ibrahim Jubran, and Dan Feldman.
\newblock Fast and accurate least-mean-squares solvers.
\newblock In {\em Advances in Neural Information Processing Systems}, pages
  8305--8316, 2019.

\bibitem{mallows2000some}
Colin~L Mallows.
\newblock Some comments on cp.
\newblock {\em Technometrics}, 42(1):87--94, 2000.

\bibitem{movckus1975bayesian}
Jonas Mo{\v{c}}kus.
\newblock On bayesian methods for seeking the extremum.
\newblock In {\em Optimization techniques IFIP technical conference}, pages
  400--404. Springer, 1975.

\bibitem{olson2016automating}
Randal~S Olson, Ryan~J Urbanowicz, Peter~C Andrews, Nicole~A Lavender, Jason~H
  Moore, et~al.
\newblock Automating biomedical data science through tree-based pipeline
  optimization.
\newblock In {\em European Conference on the Applications of Evolutionary
  Computation}, pages 123--137. Springer, 2016.

\bibitem{paszke2017automatic}
Adam Paszke, Sam Gross, Soumith Chintala, Gregory Chanan, Edward Yang, Zachary
  DeVito, Zeming Lin, Alban Desmaison, Luca Antiga, and Adam Lerer.
\newblock Automatic differentiation in pytorch.
\newblock {\em NIPS 2017}, 2017.

\bibitem{paszke2019pytorch}
Adam Paszke, Sam Gross, Francisco Massa, Adam Lerer, James Bradbury, Gregory
  Chanan, Trevor Killeen, Zeming Lin, Natalia Gimelshein, Luca Antiga, et~al.
\newblock Pytorch: An imperative style, high-performance deep learning library.
\newblock In {\em Advances in Neural Information Processing Systems}, pages
  8024--8035, 2019.

\bibitem{pedregosa2016hyperparameter}
Fabian Pedregosa.
\newblock Hyperparameter optimization with approximate gradient.
\newblock In {\em International Conference on Machine Learning}, pages
  737--746, 2016.

\bibitem{pedregosa2011scikit}
Fabian Pedregosa, Ga{\"e}l Varoquaux, Alexandre Gramfort, Vincent Michel,
  Bertrand Thirion, Olivier Grisel, Mathieu Blondel, Peter Prettenhofer, Ron
  Weiss, Vincent Dubourg, et~al.
\newblock Scikit-learn: Machine learning in python.
\newblock {\em Journal of machine learning research}, 12(Oct):2825--2830, 2011.

\bibitem{real2017large}
Esteban Real, Sherry Moore, Andrew Selle, Saurabh Saxena, Yutaka~Leon Suematsu,
  Jie Tan, Quoc~V Le, and Alexey Kurakin.
\newblock Large-scale evolution of image classifiers.
\newblock In {\em Proceedings of the 34th International Conference on Machine
  Learning-Volume 70}, pages 2902--2911. JMLR. org, 2017.

\bibitem{schmidhuber1987evolutionary}
J{\"u}rgen Schmidhuber.
\newblock {\em Evolutionary principles in self-referential learning, or on
  learning how to learn: the meta-meta-... hook}.
\newblock PhD thesis, Technische Universit{\"a}t M{\"u}nchen, 1987.

\bibitem{shahriari2015taking}
Bobak Shahriari, Kevin Swersky, Ziyu Wang, Ryan~P Adams, and Nando De~Freitas.
\newblock Taking the human out of the loop: A review of bayesian optimization.
\newblock {\em Proceedings of the IEEE}, 104(1):148--175, 2015.

\bibitem{silver2016mastering}
David Silver, Aja Huang, Chris~J Maddison, Arthur Guez, Laurent Sifre, George
  Van Den~Driessche, Julian Schrittwieser, Ioannis Antonoglou, Veda
  Panneershelvam, Marc Lanctot, et~al.
\newblock Mastering the game of go with deep neural networks and tree search.
\newblock {\em nature}, 529(7587):484, 2016.

\bibitem{simonyan2014very}
Karen Simonyan and Andrew Zisserman.
\newblock Very deep convolutional networks for large-scale image recognition.
\newblock {\em arXiv preprint arXiv:1409.1556}, 2014.

\bibitem{smith2017cyclical}
Leslie~N Smith.
\newblock Cyclical learning rates for training neural networks.
\newblock In {\em 2017 IEEE Winter Conference on Applications of Computer
  Vision (WACV)}, pages 464--472. IEEE, 2017.

\bibitem{smith2018disciplined}
Leslie~N Smith.
\newblock A disciplined approach to neural network hyper-parameters: Part
  1--learning rate, batch size, momentum, and weight decay.
\newblock {\em arXiv preprint arXiv:1803.09820}, 2018.

\bibitem{snoek2012practical}
Jasper Snoek, Hugo Larochelle, and Ryan~P Adams.
\newblock Practical bayesian optimization of machine learning algorithms.
\newblock In {\em Advances in neural information processing systems}, pages
  2951--2959, 2012.

\bibitem{stein1981estimation}
Charles~M Stein.
\newblock Estimation of the mean of a multivariate normal distribution.
\newblock {\em The annals of Statistics}, pages 1135--1151, 1981.

\bibitem{szegedy2015going}
Christian Szegedy, Wei Liu, Yangqing Jia, Pierre Sermanet, Scott Reed, Dragomir
  Anguelov, Dumitru Erhan, Vincent Vanhoucke, and Andrew Rabinovich.
\newblock Going deeper with convolutions.
\newblock In {\em Proceedings of the IEEE conference on computer vision and
  pattern recognition}, pages 1--9, 2015.

\bibitem{thompson1933likelihood}
William~R Thompson.
\newblock On the likelihood that one unknown probability exceeds another in
  view of the evidence of two samples.
\newblock {\em Biometrika}, 25(3/4):285--294, 1933.

\bibitem{tibshirani1996regression}
Robert Tibshirani.
\newblock Regression shrinkage and selection via the lasso.
\newblock {\em Journal of the Royal Statistical Society: Series B
  (Methodological)}, 58(1):267--288, 1996.

\bibitem{virtanen2020scipy}
Pauli Virtanen, Ralf Gommers, Travis~E Oliphant, Matt Haberland, Tyler Reddy,
  David Cournapeau, Evgeni Burovski, Pearu Peterson, Warren Weckesser, Jonathan
  Bright, et~al.
\newblock Scipy 1.0: fundamental algorithms for scientific computing in python.
\newblock {\em Nature methods}, 17(3):261--272, 2020.

\bibitem{xu2015block}
Yangyang Xu and Wotao Yin.
\newblock Block stochastic gradient iteration for convex and nonconvex
  optimization.
\newblock {\em SIAM Journal on Optimization}, 25(3):1686--1716, 2015.

\bibitem{zou2005regularization}
Hui Zou and Trevor Hastie.
\newblock Regularization and variable selection via the elastic net.
\newblock {\em Journal of the royal statistical society: series B (statistical
  methodology)}, 67(2):301--320, 2005.

\end{thebibliography}

\end{document}